\documentclass[10pt,journal,compsoc]{IEEEtran}

\usepackage{amsmath}
\usepackage{amssymb}
\usepackage{longtable}
\usepackage{amsthm}
\usepackage{graphicx}
\usepackage{epstopdf}
\usepackage{subfigure}
\usepackage{mathrsfs}
\usepackage{enumerate}
\usepackage{url}
\usepackage{booktabs}
\usepackage{setspace}
\usepackage{multirow}
\usepackage{marvosym}
\usepackage{subfigure}
\usepackage{tabularx}
\usepackage[numbers,sort&compress]{natbib}
\usepackage[ruled,linesnumbered]{algorithm2e}
 
\begin{document}

\title{Maximizing Mutual Information Across Feature and Topology Views for Learning Graph Representations}
\author{Xiaolong~Fan,
        Maoguo~Gong,~\IEEEmembership{Senior Member,~IEEE,}
        Yue~Wu,~\IEEEmembership{Member,~IEEE,}
        and~Hao~Li

\thanks{{Xiaolong} {Fan}, {Maoguo} {Gong} (corresponding author), and Hao {Li} are with the School of Electronic Engineering, Key Laboratory of Intelligent Perception and Image  Understanding, Ministry of Education, Xidian University, \mbox{Xi’an}, Shaanxi Province, China (e-mail: xiaolongfan@outlook.com; gong@ieee.org; haoli@xidian.edu.cn)}

\thanks{{Yue} {Wu} is with the School of Computer Science and Technology, Xidian University, \mbox{Xi’an}, Shaanxi Province, China (e-mail: ywu@xidian.edu.cn)}}

{}

\IEEEtitleabstractindextext{
\begin{abstract}
	Recently, maximizing mutual information has emerged as a powerful method for unsupervised graph representation learning. The existing methods are typically effective to capture graph information from the topology view but consistently ignore the feature view. To circumvent this issue, we propose a novel approach by exploiting mutual information maximization across feature and topology views. Specifically, we first utilize a multi-view representation learning module to better capture both local and global information content across feature and topology views on graphs. To model the information shared by the feature and topology spaces, we then develop a common representation learning module via using mutual information maximization and reconstruction loss minimization. Here, minimizing reconstruction loss forces the model to learn the shared information of feature and topology spaces. To explicitly encourage diversity between graph representations from the same view, we also introduce a disagreement regularization to enlarge the distance between representations from the same view. Experiments on synthetic and real-world datasets demonstrate the effectiveness of integrating feature and topology views. In particular, compared with the previous supervised methods, our proposed method can achieve comparable or even better performance under the unsupervised representation and linear evaluation protocol.      
\end{abstract}

}
\maketitle

\IEEEpeerreviewmaketitle

\section{Introduction}
\IEEEPARstart{W}{ith} the promising progress of Graph Neural Networks (GNNs) \cite{gnn}, supervised learning on graphs has shown outstanding performance in a number of domains, ranging from recommendation system \cite{recommend}, anomaly detection \cite{Anomaly_Detection}, fault diagnosis \cite{fault_diagnosis}, network embedding \cite{fxl}, graph Hashing \cite{hashing}, and other domains. However, supervised learning on graphs also confronts challenges such as difficulty in obtaining labeled data and generalization bias \cite{self}. To alleviate these issues, unsupervised learning methods on graphs \cite{vgae,dgi,gmi,mvdgi,you2020graph,mscsn,zeng2020contrastive,zhu2021graph} have been proposed and attracted a lot of interest. The main purpose of this method to encode topology and feature information of graph so that the learned representations can be easily utilized in the downstream tasks by using a standard machine learning algorithm. 

As a typical method of unsupervised learning on graphs, learning graph representation based on Mutual Information (MI) maximization has become a research spotlight recently. The first method, Deep Graph InfoMax (DGI) \cite{dgi}, learns the node embeddings by maximizing the mutual information between the local patch representations and a high-level global summary of the graph. Here, the local patch representation is generated through a message passing based neural network such as Graph Convolutional Network (GCN) \cite{gcn}, and the global graph summary is obtained by a readout function. Maximizing this kind of mutual information has been proved to be equivalent to maximizing the one between node latent representations and input features, which is similar to Deep InfoMax (DIM) \cite{dim}. Along this way, Contrastive Multi-View Representation (CMVR) \cite{mvdgi} on graphs is proposed to train graph encoder by maximizing mutual information between representations encoded from different topology views of graphs such as Personalized PageRank (PPR) and heat kernel of the graph. This multi-topological view learning approach has also been verified to be effective in node and graph representation learning.

Although these methods have achieved excellent performance, we observe that they only focus on the topology view to extract graph features but ignore the feature view. The recent work \cite{amgcn} has pointed out that information from node feature space also plays a crucial role in GNNs, which can enhance the representation capacity of models.  Inspired by this idea, we naturally want insight into whether maximizing mutual information by integrating features and topology structures can improve unsupervised graph representation learning. For this purpose, we try to develop our method using mutual information maximization across feature and topology views in this paper. 

Firstly, a multi-view representation module is proposed to capture local and global information content across feature and topology views. Concretely, to generate the graph of feature view, we construct the $k$-Nearest Neighbor (KNN) graph to capture the underlying structure information of nodes in feature spaces. The graph encoder can be trained by maximizing mutual information between representations encoded from both feature and topology views. We notice that the feature view and the topology view are not completely independent and some downstream tasks may require learning shared information about feature and topology spaces. To model this shared information, we then develop a common representation learning module that maximizes mutual information between common patch and common summary representations and minimizes reconstruction loss. The former follows the standard scheme of Deep Graph InfoMax (DGI) \cite{dgi} and aims to encourage the common encoder to carry the discriminative information, while the latter attempts to simultaneously reconstruct the hidden representations to the topology and feature input which encourages the common encoder to learn the shared representations of the two views. The intuitive understanding of this method is that the shared representation can maximally reconstruct the input from two views. Additionally, considering that an input graph from the same view generates two representations, we introduce the disagreement regularization to encourage diversity between graph representations generated by the same view. Here, we utilize the negative cosine similarity to measure the distance of two representations and then calculate the mean value as the final optimization objective.  

To verify the effectiveness of the developed method, we conduct extensive experiments on synthetic and real-world datasets. The experimental results on the synthetic datasets show that the previous unsupervised graph representation methods that ignore the feature view bring performance degradation compared with the simple baseline Multi-Layer Perceptron (MLP) \cite{gat}, while our proposed method achieves a significant improvement. Meanwhile, our proposed method achieve new state-of-the-art results compared with the previous unsupervised methods, and achieve comparable or even better performance under the unsupervised representation and linear evaluation protocol compared with the previous supervised methods on real-world datasets. Through experimental verification, we show that maximizing mutual information across feature and topology views can improve the performance of the unsupervised graph representation learning. 

To summarize, we outline the main contributions in this paper as below: 
\begin{itemize} 
	\item We show that integrating information from feature and topology views can improve mutual information based unsupervised graph representation learning.
	\item We propose a novel method to learn graph representations by using mutual information maximization across feature and topology views.  
	\item Experimental results on synthetic and real-world datasets indicates the effectiveness of our proposed method.    
\end{itemize}

The rest of this paper is organized as follows. In section \ref{s2}, we make a brief review of the related works of graph neural networks and contrastive self-supervised methods. Section \ref{s3} discusses the background including mutual information and estimation, and mutual information maximization on graphs. Section \ref{s4} introduces our proposed method, maximizing mutual information across feature and topology views. Section \ref{s5} evaluates the proposed method with extensive experiments on synthetic and real-world datasets. Finally, section \ref{s6} concludes this paper and discusses the future work.

\section{Related Work} \label{s2}

\subsection{Graph Neural Networks}
Graph Neural Networks (GNNs) aim to model the non-Euclidean graph data structure. As a unified framework for graph neural networks, graph message passing neural network (MPNN) \cite{mpnn,fxltnnls} generalizes several existing representative graph neural networks, such as Graph Convolutional Network (GCN) \cite{gcn}, Graph Attention Network (GAT) \cite{gat}, APPNP \cite{appnp}, and AM-GCN \cite{amgcn}, which consists of two functions: the message passing function and the readout function. For the mutual information maximization based unsupervised graph representation learning, the message passing function can be regard as the graph encoder to generate the hidden representations and the readout function can serve as the summary function to generate the summary representation vectors. In this paper, we follow the settings of DGI \cite{dgi} and choose GCN as the encoder and the mean readout as the summary function.  

\subsection{Contrastive Self-Supervised Methods}
As an important approach of self-supervised learning, Contrastive Learning (CL) aims to group similar samples closer and diverse samples far from each other. Based on this idea, several contrastive self-supervised methods \cite{cpc,moco,simclr} have been proposed for image, text, and speech data, which have achieved competitive performance, even exceeding supervised learning. The success of these methods has greatly inspired those applying contrastive learning methods for unsupervised graph representation learning, and many approaches have been proposed, such as DGI \cite{dgi}, CMVR \cite{mvdgi}, GCC \cite{mvdgi}, HDMI \cite{hdmi}, and GMI \cite{gmi}. In this paper, our developed method based on mutual information maximization actually falls under the category of contrastive learning on graphs.          

\section{Background} \label{s3}
We first introduce the notations throughout this paper. Let $G = (\mathcal{V}, \mathcal{E})$ be a graph with $\mathcal{V}$ and $\mathcal{E}$ denoting the node set and edge set, respectively. Let $n:= \lvert \mathcal{V}\rvert$ denotes the number of nodes and $\mathbf{A}$ be the $n \times n$ adjacency matrix corresponding to the graph $G$ with $\mathbf{A}(i, j) = 1$ denoting the edge between the node $v_i$ and $v_j$. The nodes are described by the feature matrix $\mathbf{X} \in \mathbb{R}^{n \times d}$ with the number of feature $d$ per node. 

\subsection{Mutual Information and Estimation}
The Mutual Information (MI) of two random variables is a measure of the mutual dependence between the two variables. Formally, MI between random variables $\boldsymbol{x}$ and $\boldsymbol{z}$ can be defined as the Kullback-Leibler (KL) divergence between the joint and the product of marginal distributions, \textit{i.e.}, 
\begin{gather}
	\mathcal{I}(\boldsymbol{x}; \boldsymbol{z})
	= \mathbb{E}_{(\boldsymbol{x}, \boldsymbol{z}) \sim p(\boldsymbol{x}, \boldsymbol{z})} \left[ \log \frac{p(\boldsymbol{x}, \boldsymbol{z})}{p(\boldsymbol{x})p(\boldsymbol{z})} \right].
\end{gather}
Intuitively, MI measures how much we can learn about the random variable $\boldsymbol{x}$ through observing $\boldsymbol{z}$.
To learn the discriminative representation $\boldsymbol{z}$, an encoder $\mathcal{E}_{\phi}(\cdot)$ is utilized to encode input $\boldsymbol{x}$ by maximizing the mutual information between $\boldsymbol{x}$ and $\boldsymbol{z}$, which can be formally defined as
\begin{gather}
	\max_{\phi} \, \mathcal{I}\left(\boldsymbol{x}; \, \mathcal{E}_{\phi}(\boldsymbol{x})\right),
\end{gather}
where $\phi$ is the learnable parameters of encoder.

However, computing this kind of mutual information directly is not feasible since the distributions are intractable. Instead, the commonly used approach is to approximate the mutual information and maximize the lower bound of the mutual information \cite{mine}. Here, we follow the intuitions from Deep InfoMax (DIM) \cite{dim} and present the objective with the Jensen-Shannon (JS) divergence mutual information estimator, \textit{i.e.},
\begin{equation}
	\begin{aligned}
		\max_{ \psi}\Big(&\mathbb{E}_{(\boldsymbol{x}, \boldsymbol{z})\sim p(\boldsymbol{x}, \boldsymbol{z})}\Big[\log \sigma\big(\mathcal{T}_{\psi}(\boldsymbol{x}, \boldsymbol{z})\big)\Big] 
		\\& + \mathbb{E}_{(\boldsymbol{x}, \boldsymbol{z})\sim p(\boldsymbol{x})p(\boldsymbol{z})}\Big[\log\big[1-\sigma\big(\mathcal{T}_{\psi}(\boldsymbol{x}, \boldsymbol{z})\big)\big]\Big]\Big),
	\end{aligned}
	\label{eq3}
\end{equation}
where $\sigma$ is a nonlinear activation, $\boldsymbol{z} = \mathcal{E}_{\phi}(\boldsymbol{x})$, and $\mathcal{T}_{\psi}(\cdot)$ is a discriminator network where $\psi$ is the learnable parameters. Equation \ref{eq3} presents a mutual information estimation approach by training the discriminator $\mathcal{T}_{\psi}(\cdot)$ between samples from the joint distribution $p(\boldsymbol{x}, \boldsymbol{z})$, \textit{i.e.}, positive samples, and samples from the product of marginals distribution $p(\boldsymbol{x})p(\boldsymbol{z})$, \textit{i.e.}, negative samples.

\subsection{Mutual Information Maximization on Graphs} 

\begin{algorithm}[t]
	\label{mmi_algo}
	\LinesNumbered
	\caption{Maximizing Mutual Information on Graphs for learning node representations}
	\KwIn{Node feature matrix $\mathbf{X}$, adjacency matrix $\mathbf{A}$, corruption function $\mathcal{C}$, encoder $\mathcal{E}_{\phi}$, summary function $\mathcal{S}$, discriminator network $\mathcal{T}_{\psi}$}
	\KwOut{Hidden representation $\left\{\boldsymbol{z}^{(1)}, \boldsymbol{z}^{(2)}, ..., \boldsymbol{z}^{(n)}\right\}$}
	\While{not converage}{
		Sampling a negative example by using 	  	corruption function: $(\tilde{\mathbf{X}}, \tilde{\mathbf{A}}) \sim \mathcal{C}\left(\mathbf{X}, \mathbf{A}\right)$\;
		Generating hidden representations for input graph by using encoder: $\mathbf{Z} = \left\{\boldsymbol{z}^{(1)}, ..., \boldsymbol{z}^{(n)}\right\} = \mathcal{E}_{\phi}\left(\mathbf{X}, \mathbf{A}\right)$\;
		Generating hidden representations for negative graph by using encoder: $\tilde{\mathbf{Z}} = \left\{\tilde{\boldsymbol{z}}^{(1)}, ..., \tilde{\boldsymbol{z}}^{(n)}\right\} = \mathcal{E}_{\phi}\left(\tilde{\mathbf{X}}, \tilde{\mathbf{A}}\right)$\;
		Obtaining summary representation by using readout function: $\boldsymbol{s} = \mathcal{S}(\mathbf{Z}) = \mathcal{S}\left(\boldsymbol{z}^{(1)}, ..., \boldsymbol{z}^{(n)}\right)$\;
		\For{$i=1$ to $n$}{
			$\mathcal{L}_{pos}^{(i)} = \mathbb{E}_{(\boldsymbol{z}^{(i)}, \boldsymbol{s})}\left[ \sigma(\mathcal{T}_{\psi}(\boldsymbol{z}^{(i)}, \boldsymbol{s}))\right]$\;
			$\mathcal{L}_{neg}^{(i)} = \mathbb{E}_{(\tilde{\boldsymbol{z}}^{(i)}, \boldsymbol{s})}\left[\log[1-\sigma(\mathcal{T}_{\psi}(\tilde{\boldsymbol{z}}^{(i)}, \boldsymbol{s}))]\right]$\;
		}
		Computing gradients: $\nabla_{\psi,\phi} \frac{1}{n}\sum_{i=1}^{n}\left(\mathcal{L}_{pos}^{(i)} + \mathcal{L}_{neg}^{(i)} \right)$\;
		Updating parameters of encoder $\mathcal{E}_{\phi}$ and discriminator $\mathcal{T}_{\psi}$ by applying stochastic gradient descent algorithm\;
	}
\end{algorithm}

Defining an appropriate mutual information objective function is critical for learning a good representation for unsupervised graph representation learning. Specifically, a graph encoder $\mathcal{E}_{\phi}(\cdot)$ is first defined so that the node representation $\boldsymbol{z}$ can be encoded using $\mathcal{E}_{\phi}(\cdot)$. The MI objective can be defined as $\max_{\phi} \mathcal{I}\left(\boldsymbol{x};\, \mathcal{E}_{\phi}(\boldsymbol{x})\right)$. Different from DIM \cite{dim} for modeling Enclidean data, the objective of DGI \cite{dgi} is maximizing mutual information between local patch representation $\boldsymbol{z}$ and summary representation $\boldsymbol{s}$, which has been proved to be equivalent to maximize mutual information between $\boldsymbol{x}$ and $\boldsymbol{z}$ in \cite{dgi}. Hence, the objective of unsupervised graph representation learning can be defined as 
\begin{gather}
	\max_{\phi}\, \frac{1}{n} \sum_{i=1}^{n} \mathcal{I}\left(\boldsymbol{z}^{(i)};\, \boldsymbol{s}\right),
\end{gather}
where $\phi$ are the parameters of encoder, $n$ is the number of nodes, $\boldsymbol{z}^{(i)}$ is the representation vector of $i$-th node, and summary representation $\boldsymbol{s}$ can be generated by summary function $\boldsymbol{s} = \mathcal{S}\left(\boldsymbol{z}^{(1)}, \boldsymbol{z}^{(2)}, ..., \boldsymbol{z}^{(n)}\right)$.  

Following Mutual Information Neural Estimation (MINE) \cite{mine}, the neural discriminator can be trained to approximate the MI and maximize the lower bound of the MI using Jensen-Shannon (JS) divergence MI estimator in form of 
\begin{equation}
	\begin{aligned}
		\max_{\phi, \psi}\bigg(\frac{1}{n} \sum_{i=1}^{n}\Big(&\mathbb{E}_{(\boldsymbol{z}^{(i)}, \boldsymbol{s})}\Big[\log \sigma\big(\mathcal{T}_{\psi}(\boldsymbol{z}^{(i)}, \boldsymbol{s})\big)\Big] 
		\\& + \mathbb{E}_{(\tilde{\boldsymbol{z}}^{(i)}, \boldsymbol{s})}\Big[\log\big[1-\sigma\big(\mathcal{T}_{\psi}(\tilde{\boldsymbol{z}}^{(i)}, \boldsymbol{s})\big)\big]\Big]\Big)\bigg),
	\end{aligned}
\end{equation} 
where $\mathcal{T}_{\psi}$ denotes a neural discriminator to provide probability scores for sampled pairs. To optimize this objective, we treat $(\boldsymbol{z}^{(i)}, \boldsymbol{s})$ as positive samples, sampled from the joint distribution. To obtain the samples from the product of marginal distribution, we can use the corruption function $\mathcal{C}(\cdot)$ to generate the negative samples $\tilde{\boldsymbol{z}}^{(i)}$. With $\tilde{\boldsymbol{z}}^{(i)}$ and $\boldsymbol{s}$, the negative samples can be regarded as samples sampled from the product of marginal distribution. Therefore, the objective of MI can be easily optimized using gradient descent algorithm. For clarity, we present the overall algorithm scheme in Algorithm \ref{mmi_algo}.

\begin{figure*}[]
	\includegraphics[width=1.97\columnwidth]{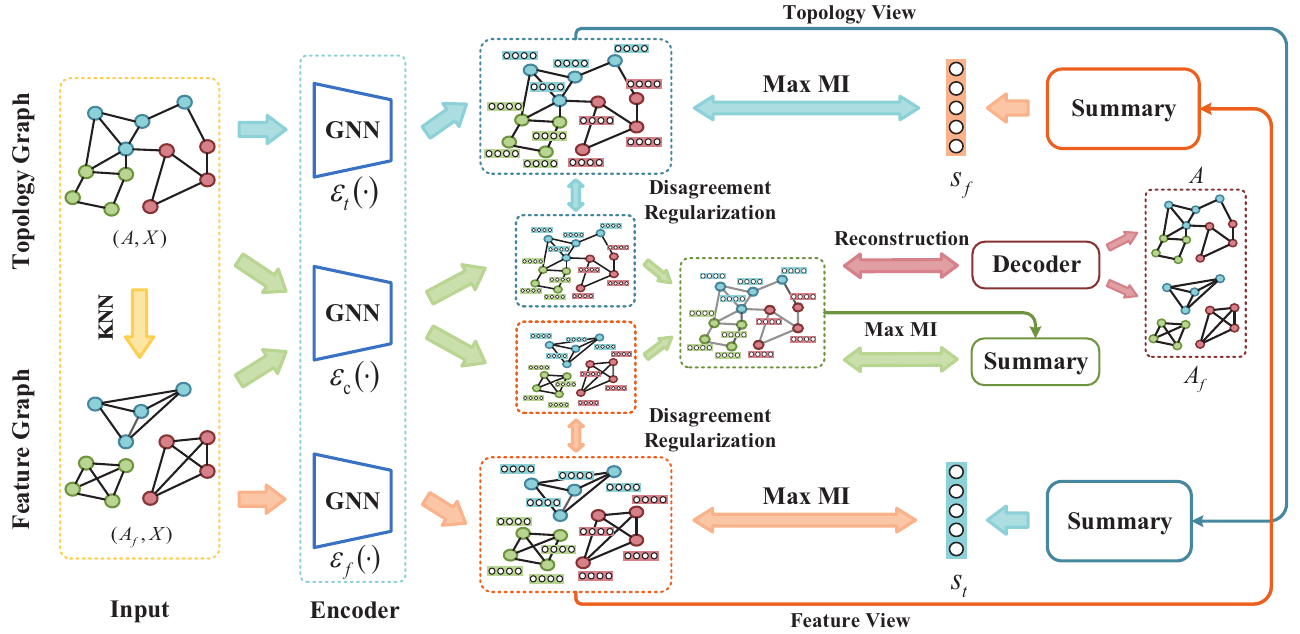}
	\centering
	\caption{The overall framework of the developed method. The multi-view representation learning module is utilized to better capture both local and global information content across feature and topology views on graphs. The common representation learning module using mutual information maximization and reconstruction loss minimization is utilized to model the information shared by the feature and topology spaces. And, the disagreement regularization is employed to enlarge the distance between representations from the same view.}
	\label{framework}
\end{figure*}

\section{Methodology} \label{s4}
In this section, we will present our proposed method to learn node representations across feature and topology views, which consists of multi-view Representation learning module, common representation learning module, and disagree regularization. The overall framework is shown in Figure \ref{framework}.

\subsection{Generating Feature Graph}
To generate the feature graph, we construct the $k$-Nearest Neighbor (KNN) graph to capture the underlying structure of nodes in feature spaces. Specifically, cosine distance is utilized to find the similarity matrix $\mathbf{S} \in \mathbb{R}^{n\times n}$ among $n$ nodes, \textit{i.e.}, 
\begin{gather}
	\mathbf{S}_{i,j} = \frac{\boldsymbol{x}^{(i)} \cdot \boldsymbol{x}^{(j)}}{\Vert \boldsymbol{x}^{(i)} \Vert \cdot \Vert \boldsymbol{x}^{(j)} \Vert} \label{eq5}.
\end{gather}

After obtaining the similarity matrix $\mathbf{S}$, we can choose top $K$ similar node pairs for each node to form their neighbors and finally generate the feature graph $G_f = \left(\mathbf{A}_f, \mathbf{X}\right)$ from the original graph $G = \left(\mathbf{A}, \mathbf{X}\right)$. Note that the adjacency relationship of feature graph $G_f$ is only determined by node features. 

\subsection{Multi-View Representation Module}
Given two views of input graph data, \textit{i.e.}, feature view $G_f = (\mathbf{A}_f, \mathbf{X})$ and topology view $G = (\mathbf{A}, \mathbf{X})$, we can utilize the standard multi-view contrastive representation learning framework \cite{mvdgi} to generate node representations. Firstly, two GCN encoders are used to generate feature patch representations $\mathbf{Z}_f$ and topology patch representations $\mathbf{Z}_t$, respectively, \textit{i.e.},
\begin{gather}
	\mathbf{Z}_f = \mathcal{E}_f(\mathbf{X}, \mathbf{A}_f) = \sigma\left(\mathbf{\hat{D}}^{-\frac{1}{2}}_f \mathbf{\hat{A}}_f
	\mathbf{\hat{D}}^{-\frac{1}{2}}_f \mathbf{X} \mathbf{\Theta}_f\right), \label{eq6}\\
	\mathbf{Z}_t = \mathcal{E}_t(\mathbf{X}, \mathbf{A}) = \sigma\left(\mathbf{\hat{D}}^{-\frac{1}{2}}\mathbf{\hat{A}}
	\mathbf{\hat{D}}^{-\frac{1}{2}} \mathbf{X} \mathbf{\Theta}_t\right) \label{eq7},
\end{gather} 
where $\mathbf{\hat{A}}_f = \mathbf{A}_f + \mathbf{I}$, $\mathbf{\hat{A}} = \mathbf{A} + \mathbf{I}$, $\mathbf{\hat{D}}_f$ and $\mathbf{\hat{D}}$ are the degree matrix, $\mathbf{\hat{D}} = diag(\hat{\boldsymbol{d}})$, where $\hat{\boldsymbol{d}} \in \mathbb{R}^{n}$ is the degree vector with $\hat{\boldsymbol{d}_j} = \sum_{i}\hat{\mathbf{A}}_{ij}$, $\mathbf{\Theta}_f$ and $\mathbf{\Theta}_t$ are the learnable parameters. To capture the global information content, we then leverage the mean function as readout to summarize the patch representations of two views in form of 
\begin{gather}
	\boldsymbol{s}_f = \mathcal{S}_f(\mathbf{Z}_f) = \mathcal{S}_f\left(\boldsymbol{z}_f^{(1)}, \boldsymbol{z}_f^{(2)}, ..., \boldsymbol{z}_f^{(n)}\right) =  \sigma\left(\frac{1}{n}\sum_{i=1}^{n}\boldsymbol{z}_f^{(i)}\right)\label{eq8}, \\
	\boldsymbol{s}_t = \mathcal{S}_t(\mathbf{Z}_t) = \mathcal{S}_t\left(\boldsymbol{z}_t^{(1)}, \boldsymbol{z}_t^{(2)}, ..., \boldsymbol{z}_t^{(n)}\right) =  \sigma\left(\frac{1}{n}\sum_{i=1}^{n}\boldsymbol{z}_t^{(i)}\right)\label{eq9}.
\end{gather} 
With the patch and summary representations, we can organize the objective of the multi-view representation module as follows
\begin{gather}
	\mathcal{L}_{mmi} = \frac{1}{n} \sum_{i=1}^{n}\left[\mathcal{I}\big(\boldsymbol{z}_f^{(i)};\, \boldsymbol{s}_t\big) + \mathcal{I}\big(\boldsymbol{z}_t^{(i)}; \, \boldsymbol{s}_f\big)\right],
	\label{mvloss}
\end{gather}   
where $n$ is the number of nodes, $\boldsymbol{z}_f^{(i)}$ and $\boldsymbol{z}_t^{(i)}$ denote the hidden representations of feature space and topology space, respectively. 

\subsection{Common Representation Module}
Note that the valid node representations may be related to both the feature and the topology spaces. Therefore, an additional module needs to be developed to capture the common information shared by these two spaces. To this end, we utilize a weight-sharing GCN encoder to extract two common representations across two spaces and then concatenate these two representations to generate the common representation, \textit{i.e.}, 
\begin{gather}
	\mathbf{Z}_{cf} = \mathcal{E}_c(\mathbf{X}, \mathbf{A}_f) = \sigma\left(\mathbf{\hat{D}}^{-\frac{1}{2}}_f \mathbf{\hat{A}}_f
	\mathbf{\hat{D}}^{-\frac{1}{2}}_f \mathbf{X} \mathbf{\Theta}_c\right)\label{eq11}, \\
	\mathbf{Z}_{ct} = \mathcal{E}_c\left(\mathbf{X}, \mathbf{A}\right) = \sigma\left(\mathbf{\hat{D}}^{-\frac{1}{2}}\mathbf{\hat{A}}
	\mathbf{\hat{D}}^{-\frac{1}{2}} \mathbf{X} \mathbf{\Theta}_c\right)\label{eq12}, \\
	\mathbf{Z}_{c} = \textsc{Mlp}\left(\mathbf{Z}_{cf}\,\Vert\,\mathbf{Z}_{ct}\right)\label{eq13},
\end{gather}
where $\textsc{Mlp}(\cdot)$ represents the Multi-Layer Perceptron and $\Vert$ denotes the concatenation operator. The summary representation $\boldsymbol{s}_c$ can be generated by using mean readout function, \textit{i.e.}, 
\begin{gather}
	\boldsymbol{s}_c = \mathcal{S}_c\left(\mathbf{Z}_c\right) = \mathcal{S}_c\left(\boldsymbol{z}_c^{(1)}, \boldsymbol{z}_c^{(2)}, ..., \boldsymbol{z}_c^{(n)}\right) = \sigma\left(\frac{1}{n}\sum_{i=1}^{n}\boldsymbol{z}_c^{(i)}\right)\label{eq14}.
\end{gather}
Analogously, we optimize the mutual information between common representations $\mathbf{Z}_{c} = \{\boldsymbol{z}^{(1)}, \boldsymbol{z}^{(2)}, ..., \boldsymbol{z}^{(N)}\}$ and summary representation $\boldsymbol{s}_c$ in form of 
\begin{gather}
	\mathcal{L}_{cmi} = \frac{1}{n} \sum_{i=1}^{n} \mathcal{I}\left(\boldsymbol{z}_c^{(i)};\, \boldsymbol{s}_c\right).
	\label{closs}
\end{gather}

Considering our target is to learn common representations across two spaces, we simultaneously reconstruct the output of common encoder to feature and topology spaces to learn common node representations in an encode-decode paradigm. In this way, the model tends to learn the shared information of feature and topology spaces. Specifically, the decoder is a inner product \cite{vgae} of the representations for reconstructing the adjacency matrix, \textit{i.e.}, 
\begin{gather}
	\mathbf{A}^{'}_f = \sigma\left(\mathbf{Z}_{c}\mathbf{Z}_{c}^{\top}\right)\label{eq16}, \\
	\mathbf{A}^{'} = \sigma\left(\mathbf{Z}_{c}\mathbf{Z}_{c}^{\top}\right)\label{eq17},
\end{gather}
where $\sigma$ is the Sigmoid function, $\mathbf{A}^{'}_f$ and $\mathbf{A}^{'}$ are the reconstructed adjacency matrixes. The objective is to minimize the reconstruction loss between the input adjacency matrix and reconstructed adjacency matrix. To optimize the reconstruction loss, we employ the contrastive learning approach by randomly sampling positive and negative nodes from neighbors and non-neighbors respectively. Therefore, the reconstruction objective can be converted to the cross-entropy loss of positive and negative node pairs, and we can minimize this loss, \textit{i.e.},
\begin{equation}
	\begin{aligned}
		\mathcal{L}_r = &\left[-\log(\mathbf{A}^{'}_{fpos}) - \log(1 - \mathbf{A}^{'}_{fneg})\right] \\ 
		&+ \left[-\log(\mathbf{A}^{'}_{pos}) - \log(1 - \mathbf{A}^{'}_{neg})\right],
		\label{recontloss}
	\end{aligned} 
\end{equation}
where $\mathbf{A}^{'}_{fpos}$ and $\mathbf{A}^{'}_{pos}$ are the adjacency matrixes formed by positive node pairs, \textit{i.e.}, nodes from neighbors, $\mathbf{A}^{'}_{fneg}$ and $\mathbf{A}^{'}_{neg}$ formed by negative node pairs, \textit{i.e.}, nodes randomly from non-neighbors.   

\subsection{Disagreement Regularization}
Note that an input graph from the same view generates two feature representations, \textit{i.e.}, $(\mathbf{A}_f, \mathbf{X}_f) \to$ $\mathbf{Z}_f, \, \mathbf{Z}_{cf}$ and $(\mathbf{A}, \mathbf{X}) \to \mathbf{Z}_t, \, \mathbf{Z}_{ct}$. To explicitly encourage the diversity between representations from same view, we introduce a disagreement regularization to enlarge the distances, inspired by \cite{multi}. Specifically, we utilize the negative cosine similarity to measure the distance of two vectors and then calculate the mean value as the final measurement, \textit{i.e.}, 
\begin{gather}
	\mathcal{L}_d = -\frac{1}{n}\sum_{i=1}^{n} \left[ \frac{\boldsymbol{z}_f^{(i)} \cdot \boldsymbol{z}_{cf}^{(i)}}{\Vert \boldsymbol{z}_f^{(i)} \Vert \cdot \Vert \boldsymbol{z}_{cf}^{(i)} \Vert} + \frac{\boldsymbol{z}_t^{(i)} \cdot \boldsymbol{z}_{ct}^{(i)}}{\Vert \boldsymbol{z}_t^{(i)} \Vert \cdot \Vert \boldsymbol{z}_{ct}^{(i)} \Vert}\right].
	\label{cosloss} 
\end{gather}
Our learning objective is to enlarge $\mathcal{L}_d$ to guarantee the diversity in unsupervised graph representation learning.

\subsection{Optimization and Inference}
In summary, the final objective function of the developed model consists of four parts: (1) the MI objective for multi-view representation learning (Equation \ref{mvloss}); (2) the MI objective for common representation learning (Equation \ref{closs}); (3) the reconstruction objective for common representation learning (Equation \ref{recontloss}); (4) the disagreement regularization objective (Equation \ref{cosloss}). Considering that the objectives of MI and disagreement regularization need to be maximized and reconstruction to be minimized, we rewrite the objective as follows
\begin{gather}
	\mathcal{L} = -\left[\mathcal{L}_{mmi} + \lambda_c \cdot (\mathcal{L}_{cmi} - \mathcal{L}_{r}) + \lambda_d \cdot \mathcal{L}_d\right]\label{allloss},
\end{gather}
where $\lambda_c$ and $\lambda_d$ are the trade-off parameters of common representation module objective $(\mathcal{L}_{cmi} - \mathcal{L}_{r})$ and disagreement regularization objective $\mathcal{L}_d$, respectively. We can easily use the standard gradient descent algorithm to minimize this objective. The whole training process is shown in Algorithm \ref{all_algo}.

At inference time, the parameter-free mean function is employed to aggregate representations generated by multi-view and common modules, \textit{i.e.},
$\mathbf{Z} = \textsc{Mean}\big(\mathbf{Z}_f, \mathbf{Z}_t, \mathbf{Z}_{c}\big)$, for downstream tasks such as classification and clustering.

\begin{algorithm}[t]
	\label{all_algo}
	\LinesNumbered
	\caption{Training process of the proposed method}
	\KwIn{Node feature matrix $\mathbf{X}$, adjacency matrix $\mathbf{A}$, neighbor number $K$, trade-off parameters $\lambda_c$ and $\lambda_d$}
	\KwOut{Hidden representation $\left\{\boldsymbol{z}^{(1)}, \boldsymbol{z}^{(2)}, ..., \boldsymbol{z}^{(n)}\right\}$}
	Generating similarity matrix $\mathbf{S}$ via Equation \ref{eq5}\;
	Generating feature graph $G_f = (\mathbf{A}_f, X)$ via choosing Top $K$ similar node pairs for each node\;
	\While{not converage}{
		Obtaining multi-view representations $\mathbf{Z}_f$ and $\mathbf{Z}_t$ via Equation \ref{eq6} and \ref{eq7}\;
		Obtaining multi-view summary vectors $\boldsymbol{s}_f$ and $\boldsymbol{s}_t$ via Equation \ref{eq8} and \ref{eq9}\;
		Obtaining common representations $\mathbf{Z}_c$ via Equation \ref{eq11}, \ref{eq12}, and \ref{eq13}\;
		Obtaining common summary vectors $\boldsymbol{s}_c$ via Equation \ref{eq14}\;
		Obtaining postive reconstructed adjacency matrix $\mathbf{A}^{'}_{fpos}$ and $\mathbf{A}^{'}_{pos}$ via sampling postive node pairs\;
		Obtaining negative reconstructed adjacency matrix $\mathbf{A}^{'}_{fneg}$ and $\mathbf{A}^{'}_{neg}$ via sampling postive node pairs\;
		Calculating the multi-view MI objective $\mathcal{L}_{mmi}$ via Equation $\ref{mvloss}$\;
		Calculating the common-view MI objective $\mathcal{L}_{cmi}$ via Equation $\ref{closs}$\;
		Calculating the reconstruction objective $\mathcal{L}_{r}$ via Equation $\ref{recontloss}$\;
		Calculating the disagreement regularization objective $\mathcal{L}_{d}$ via Equation $\ref{cosloss}$\;
		Calculating the overall objective $\mathcal{L}$ via Equation $\ref{allloss}$\;
		Computing gradients $\nabla{\mathcal{L}}$ and updating parameters by applying stochastic gradient descent algorithm\;
	}
\end{algorithm}


\section{Experiment} \label{s5}

\subsection{Datasets}
We conduct comparative experiments on two synthetic datasets including \textsc{Fea. Syn. Data} and \textsc{Top. Syn. Data}, and five real-world datasets including \textsc{Cora}, \textsc{CiteSeer}, \textsc{PubMed}, \textsc{PubMedFull}, \textsc{Amazon Computers}, and \textsc{Amazon Photo}. The statistics of all datasets are shown in Table \ref{tabd}.
\subsubsection{Synthetic Datasets}

\begin{table}[b]
	\caption{Statistics of two synthetic and five real-world datasets. Ama. Computers and Ama. Photo represent Amazon Computers and Amazon Photo datasets, respectively.}
	\scalebox{0.8}{
		\begin{tabular}{|c|c|c|c|c|c|}
			\toprule
			&\textsc{Datasets}&\textsc{Nodes}&\textsc{Edges}&\textsc{Features}&\textsc{Classes}\\
			\midrule
			\midrule
			\multirow{2}*{\rotatebox{90}{\textsc{Syn.}}}&\textsc{Fea. Syn.}&2,400&59,870&20&3\\
			&\textsc{Top. Syn.}&2,400&59,976&20&3\\
			\midrule
			\multirow{6}*{\rotatebox{90}{\textsc{Real-World}}}&\textsc{Cora}&2,708&5,429&1,433&7\\
			&\textsc{CiteSeer}&3,327&4,732&3,703&6\\
			&\textsc{PubMed}&19,717&44,338&500&3\\
			&\textsc{PubMedFull}&18,230&79,612&500&3\\
			&\textsc{Ama. Computers}&13,381&245,778&767&10\\
			&\textsc{Ama. Photo}&7,487&119,043&745&8\\
			\bottomrule
	\end{tabular}}
	\centering
	\label{tabd}
\end{table}

To verify the effectiveness of integrating information from feature space, we generate two random networks for evaluation. The first synthetic network, Feature Synthetic (\textsc{Fea. Syn.}), consists of $2,400$ nodes, and the probability of building an edge between any two nodes is set to $0.01$. By sampling the Gaussian distribution with the same covariance matrix but with different centers, we can obtain the feature matrix. Here, nodes with the same center distribution are assigned the same label. The second synthetic network, Topology Synthetic (\textsc{Top. Syn.}), is divided into three communities, where the probability of building edges between nodes in each community is $0.03$ and the probability of edges between any communities is $0.0015$. Similar to the first network, the node features are generated by three different Gaussian distributions. Note that different from \cite{amgcn}, nodes belonging to the same community have the same feature distribution and the same label in our settings. In general, the node label of the first network is only related to node features but not topology structures, while the second network is not only related to node features but also topology structures. Both synthetic data sets, \textsc{Syn. Fea. Data} and \textsc{Syn. Top. Data}, are divided into three classes, each of which has $800$ nodes. In evaluation phase, we randomly select $20$ nodes per class for training and other nodes for testing.

\subsubsection{Real-World Datasets} 
We use five real-world datasets, \textsc{Cora}, \textsc{CiteSeer}, \textsc{PubMed} from \cite{pladata}, \textsc{PubMedFull} from \cite{bojchevski2018deep}, \textsc{Amazon Computers} and \textsc{Amazon Photo} from \cite{shchur2018pitfalls}, for evaluation. For the citation network datasets, the datasets contain sparse bag-of-words feature vectors for each document and a list of citation links between documents. The citation links are treated as undirected edges and we can construct a binary, symmetric adjacency matrix. Each document has a class label. \textsc{Amazon Computers} and \textsc{Amazon Photo}  are segments of the Amazon co-purchase graph where nodes represent goods, edges indicate that two goods are frequently bought together, node features are bag-of-words encoded product reviews, and class labels are given by the product category. For \textsc{Cora}, \textsc{CiteSeer}, and \textsc{PubMed} datasets, we use the public dataset split where $20$ samples per label is used for training in the downstream classification task, and the predictive power of the learned representations is evaluated on $1000$ test nodes. For \textsc{PubMedFull}, \textsc{Amazon Computers}, and \textsc{Amazon Photo} datasets, we randomly select $20$ nodes per class for training and other nodes for testing.

\begin{table*}[htbp]
	\caption{Experimental results on two synthetic datasets. We report the average classification accuracy $\pm$ standard deviation. In this table, \textsc{Fea.} and \textsc{Top.} denote Feature and Topology respectively.}
	\scalebox{0.88}{
		\begin{tabular}{|c|c|c|c|c|c|}
			\toprule
			&\textsc{Methods}&\textsc{Fea. View}&\textsc{Top. View}&\textsc{Fea. Syn. Data}@\textsc{Accuracy}&\textsc{Top. Syn. Data}@\textsc{Accuracy}\\
			\midrule
			\midrule
			\multirow{4}*{\rotatebox{90}{\textsc{Sup.}}}&MLP&${\surd}$&${\times}$&\textbf{92.52 $\pm$ 0.52\%}&95.13 $\pm$ 0.21\%\\ 
			&GCN&${\times}$&${\surd}$&64.53 $\pm$ 0.33\%&\textbf{100.00\%}\\
			&GAT&${\times}$&${\surd}$&65.90 $\pm$ 0.45\%&\textbf{100.00\%}\\
			&\textsc{MixHop}&${\times}$&${\surd}$&68.41 $\pm$ 0.55\%&\textbf{100.00\%}\\
			\midrule
			\multirow{5}*{\rotatebox{90}{\textsc{Unsup.}}}&\textsc{DeepWalk}&${\times}$&${\surd}$&34.98 $\pm$ 0.32\%&\textbf{100.00\%}\\
			&VGAE&${\times}$&${\surd}$&66.15 $\pm$ 0.79\%&\textbf{100.00\%}\\
			&DGI&${\times}$&${\surd}$&38.12 $\pm$ 0.26\%&\textbf{100.00\%}\\ 
			&CMVR&${\times}$&${\surd}$&89.39 $\pm$ 0.39\%&\textbf{100.00\%}\\
			&MVMI-FT&${\surd}$&${\surd}$&\textbf{94.33 $\pm$ 0.69\%}&\textbf{100.00\%}\\
			\bottomrule
	\end{tabular}}
	\centering
	\label{tabs}
\end{table*}

\begin{table*}[htbp]
	\caption{Node classification experimental results on five real-world datasets. After 50 experimental runs, we report the average classification accuracy $\pm$ standard deviation. In this table, Ama. represents Amazon.}
	\scalebox{0.9}{
		\begin{tabular}{|c|c|c|c|c|c|c|c|c|}
			\toprule
			&\textsc{Methods}&\textsc{Input}&\textsc{Cora}&\textsc{CiteSeer}&\textsc{PubMed}&\textsc{PubMedFull}&\textsc{Ama. Computers}&\textsc{Ama. Photo}\\
			\midrule
			\midrule
			\multirow{7}*{\rotatebox{90}{\textsc{Supervised}}}&MLP&$\boldsymbol{X}$, $\boldsymbol{Y}$& 55.1 $\pm$ 1.1\%&46.5 $\pm$ 1.3\%&71.4 $\pm$ 1.1\%&70.8 $\pm$ 1.2\%&44.9 $\pm$ 1.8\%&65.6 $\pm$ 2.3\%\\ 
			&LP&$\boldsymbol{A}$, $\boldsymbol{Y}$&68.0 $\pm$ 1.6\%&63.7 $\pm$ 2.1\%&70.5 $\pm$ 1.9\%&70.0 $\pm$ 2.1\%&63.8 $\pm$ 1.9\%&72.6 $\pm$ 2.7\%\\
			&GCN&$\boldsymbol{X}$, $\boldsymbol{A}$, $\boldsymbol{Y}$&81.4 $\pm$ 0.3\%&70.3 $\pm$ 0.4\%&78.2 $\pm$ 0.6\%&79.1 $\pm$ 0.6\%&75.2 $\pm$ 1.3\%&\textbf{86.7 $\pm$ 0.8\%}\\
			&SAGE&$\boldsymbol{X}$, $\boldsymbol{A}$, $\boldsymbol{Y}$&80.9 $\pm$ 0.4\%&70.4 $\pm$ 0.7\%&77.2 $\pm$ 0.2\%&77.8 $\pm$ 0.4\%&76.8 $\pm$ 0.4\%&85.5 $\pm$ 0.3\%\\
			&GAT&$\boldsymbol{X}$, $\boldsymbol{A}$, $\boldsymbol{Y}$&82.8 $\pm$ 0.6\%&71.7 $\pm$ 0.5\% &78.6 $\pm$ 0.3\%&80.4 $\pm$ 0.5\%&\textbf{78.7 $\pm$ 1.4\%}&\textbf{86.8 $\pm$ 2.3\%}\\
			&\textsc{MixHop}&$\boldsymbol{X}$, $\boldsymbol{A}$, $\boldsymbol{Y}$&81.9 $\pm$ 0.4\%&71.4 $\pm$ 0.8\%&80.8 $\pm$ 0.6\%&79.8 $\pm$ 0.8\%&\textbf{77.8 $\pm$ 0.9\%}&\textbf{87.3 $\pm$ 1.1\%}\\
			&APPNP&$\boldsymbol{X}$, $\boldsymbol{A}$, $\boldsymbol{Y}$&\textbf{83.0 $\pm$ 0.6\%}&72.8 $\pm$ 0.3\%&80.7 $\pm$ 0.3\%&80.8 $\pm$ 0.6\%&\textbf{78.9 $\pm$ 0.7\%}&\textbf{88.1 $\pm$ 0.8\%}\\
			\midrule
			\multirow{13}*{\rotatebox{90}{\textsc{Unsupervised}}}&\textsc{Linear}&$\boldsymbol{X}$&47.9 $\pm$ 0.4\%&49.3 $\pm$ 0.2\%& 69.1 $\pm$ 0.3\%&68.3 $\pm$ 0.5\%&41.1 $\pm$ 0.7\%&52.6 $\pm$ 0.6\%\\
			&\textsc{DeepWalk}&$\boldsymbol{X}$, $\boldsymbol{A}$&69.5 $\pm$ 0.4\%&50.8 $\pm$ 0.4\%&73.1 $\pm$ 0.6\%&69.4 $\pm$ 0.3\%&60.7 $\pm$ 0.8\%&58.9 $\pm$ 0.9\%\\
			&VGAE&$\boldsymbol{X}$, $\boldsymbol{A}$&71.5 $\pm$ 0.5\%&65.6 $\pm$ 0.3\%&72.1 $\pm$ 0.5\%&71.5 $\pm$ 0.8\%&67.4 $\pm$ 1.3\%&69.9 $\pm$ 1.1\%\\
			&VGAE-S&$\boldsymbol{X}$, $\boldsymbol{A}$&72.5 $\pm$ 0.9\%&66.7 $\pm$ 0.6\%&70.1 $\pm$ 0.8\%&71.7 $\pm$ 1.1\%&65.8 $\pm$ 1.1\%&67.6 $\pm$ 1.5\%\\
			&DGI&$\boldsymbol{X}$, $\boldsymbol{A}$&82.2 $\pm$ 0.4\%&70.9 $\pm$ 0.6\%&76.6 $\pm$ 0.5\%&75.9 $\pm$ 0.9\%&74.8 $\pm$ 1.0\%&83.1 $\pm$ 0.9\%\\
			&DGI-S&$\boldsymbol{X}$, $\boldsymbol{A}$&82.2 $\pm$ 0.4\%&70.9 $\pm$ 0.6\%&76.6 $\pm$ 0.5\%&74.3 $\pm$ 1.1\%&74.1 $\pm$ 0.9\%&82.4 $\pm$ 0.5\%\\
			&DGI-D&$\boldsymbol{X}$, $\boldsymbol{A}$&\textbf{83.1 $\pm$ 0.6\%}&71.2 $\pm$ 0.6\%&76.9 $\pm$ 0.4\%&76.3 $\pm$ 0.7\%&75.3 $\pm$ 1.1\%&82.5 $\pm$ 0.8\%\\
			&CMVR&$\boldsymbol{X}$, $\boldsymbol{A}$&\textbf{83.5 $\pm$ 0.4\%}&71.6 $\pm$ 0.3\%&77.4 $\pm$ 0.3\%&77.3 $\pm$ 0.9\%&76.6 $\pm$ 0.8\%&83.2 $\pm$ 0.9\%\\
			&CMVR-S&$\boldsymbol{X}$, $\boldsymbol{A}$&\textbf{83.7 $\pm$ 0.5\%}&69.8 $\pm$ 0.6\%&76.1 $\pm$ 0.5\%&76.5 $\pm$ 0.6\%&76.4 $\pm$ 0.6\%&83.4 $\pm$ 1.2\%\\
			&MI-FT &$\boldsymbol{X}$, $\boldsymbol{A}$&82.6 $\pm$ 0.5\%&71.8 $\pm$ 0.7\%&80.8 $\pm$ 0.5\%&78.6 $\pm$ 1.4\%&75.6 $\pm$ 0.6\%&83.8 $\pm$ 1.2\%\\
			&MVMI-FT&$\boldsymbol{X}$, $\boldsymbol{A}$&\textbf{83.1 $\pm$ 0.6\%}&72.7 $\pm$ 0.5\%&\textbf{81.0 $\pm$ 0.3\%}&\textbf{81.1 $\pm$ 0.3\%}&\textbf{77.5 $\pm$ 0.7\%}	&\textbf{86.1 $\pm$ 0.2\%}\\
			&MVMI-FT-S&$\boldsymbol{X}$, $\boldsymbol{A}$&82.6 $\pm$ 0.4\%&71.6 $\pm$ 0.3\%&\textbf{81.1 $\pm$ 0.4\%}&80.7 $\pm$ 0.5\%&\textbf{77.0 $\pm$ 0.9\%}	&85.2 $\pm$ 0.7\%\\	
			&MVMI-FT-D&$\boldsymbol{X}$, $\boldsymbol{A}$&\textbf{83.8 $\pm$ 0.4\%}&\textbf{73.1 $\pm$ 0.4\%}&\textbf{81.6 $\pm$ 0.3\%}&\textbf{81.8 $\pm$ 0.4\%}&\textbf{77.8 $\pm$ 0.6\%}&\textbf{86.5 $\pm$ 0.4\%}\\	
			\bottomrule
	\end{tabular}}
	\centering
	\label{tabr}
\end{table*}

\subsection{Baselines}
We compare the proposed method with previous supervised and unsupervised state-of-the-art methods. The supervised models include: Multi-Layer Perceptron (MLP) \cite{gat}, Label Propagation (LP) \cite{lp}, Graph Convolutional Network (GCN) \cite{gcn}, Graph Sample and Aggregate Network (SAGE) \cite{sage}, Graph Attention Network (GAT) \cite{gat}, Higher-Order Graph Convolutional Architectures (MixHop) \cite{mixhop}, and Approximate Personalized Propagation of Neural Predictions (APPNP) \cite{appnp}. The unsupervised methods include: Linear \cite{dgi}, DeepWalk \cite{deepwalk}, Variational Graph Auto-Encoder (VGAE) \cite{vgae}, VGAE with SAGE encoder variant (VGAE-S), Deep Graph Infomax (DGI) \cite{dgi}, DGI with SAGE encoder variant (DGI-S), DGI with PPR Diffusion variant (DGI-D) \cite{mvdgi}, Contrastive Multi-View Representation Learning on Graphs (CMVR) \cite{mvdgi}, CMVR with SAGE encoder variant (CMVR-S).

\subsection{Experimental Configurations}
For experiments on all synthetic and real-world datasets, we set the learning rate to be $0.001$, hidden dimension to be $512$, maximum number of epochs to be $500$. We initialize the weight parameters using Glorot initialization \cite{glorot} and the bias parameters using zero initialization. All models are trained with Adam \cite{adam} optimizer. For the nonlinear activation, we use the Parametric ReLU (PReLU) \cite{prelu} as $\sigma$ in encoder and Sigmoid in summary and neural discriminator. Following DGI \cite{dgi}, we set the neural discriminator to be a simple bilinear function, and for the corruption, we preserve the original adjacency matrix $\tilde{\mathbf{A}} = \mathbf{A}$ but corrupt the features $\mathbf{X}$ by using row-wise random shuffling to generate the negative sample $\tilde{\mathbf{X}}$. Hence the number of positive samples is equal to the negative samples. We set $\lambda_c = 0.3$ and $\lambda_d = 0.01$ in final objective. For the node classification task, we conduct $50$ experimental runs and report the average accuracy with standard deviation on the test set using a linear Logistic Regression classification model with L-BFGS optimization method. For the node clustering task, we cluster the learned representations using K-Means algorithm with default settings, and we set the number of clusters to be the ground-truth classes and report the Normalized Mutual Information (NMI) and Adjusted Rand Index (ARI) scores. For all baselines, we follow the original model and experiment settings to achieve a fair comparison. All codes in this paper are implemented with PyTorch Geometric and are released in this site\footnote{\url{https://github.com/xiaolongo/MaxMIAcrossFT}}. 

\begin{table*}[t]
	\caption{Node clustering comparison on real-world datasets using normalized MI (NMI) and adjusted rand index (ARI) measures.}
	\scalebox{0.83}{
		\begin{tabular}{|c|cc|cc|cc|cc|cc|cc|}
			\toprule
			\multirow{2}{*}{\textsc{Methods}}&\multicolumn{2}{c|}{\textsc{Cora}} &
			\multicolumn{2}{c|}{\textsc{CiteSeer}} & \multicolumn{2}{c|}{\textsc{PubMed}} &
			\multicolumn{2}{c|}{\textsc{PubMedFull}} & \multicolumn{2}{c|}{\textsc{Ama. Computers}} & \multicolumn{2}{c|}{\textsc{Ama. Photo}}\\
			&NMI&ARI&NMI&ARI&NMI&ARI&NMI&ARI&NMI&ARI&NMI&ARI\\
			\midrule
			\midrule
			\textsc{Dgi}&0.5341&0.4435&0.3943&0.3842&0.2378&0.2071&0.2611&0.2366&0.3315&0.1965&0.3744&0.2445\\
			\textsc{Cmvr}&0.5471&0.4799&0.4008&0.3883&0.3305&0.2996&0.2705&0.2411&0.3376&0.1793&0.3769&0.2584\\
			\textsc{Mvmi-Ft}&\textbf{0.5524}&\textbf{0.4873}&\textbf{0.4031}&\textbf{0.3924}&\textbf{0.3385}&\textbf{0.3093}&\textbf{0.3194}&\textbf{0.2975}&\textbf{0.3818}&\textbf{0.2166}&\textbf{0.4265}&\textbf{0.2918}\\
			\midrule
			\textsc{Gain}&0.96\%&1.54\%&0.57\%&1.06\%&2.42\%&3.24\%&18.1\%&23.4\%&13.1\%&10.2\%&13.2\%&12.9\%\\
			\bottomrule
	\end{tabular}}
	\centering
	\label{tabc}
\end{table*} 

\subsection{Results on Synthetic Datasets}
The experimental results on synthetic datasets are reported  in Table \ref{tabs}.
For feature synthetic dataset (\textsc{Fea. Syn. Data}), we first compare MLP and the existing baselines, and can observe that MLP that only utilize information from feature view obtains the best classification results $92.52$ $\pm$ $0.52\%$. Conversely, DGI only achieves an accuracy of $38.12$ $\pm$ $0.26\%$, and although CMVR achieves a significant improvement, it still has a $3.13\%$ performance degradation compared with MLP. Our interpretation is that these baseline models ignore the information content from feature view. When the node label is independent of the topology, this method of aggregating information in the topology space introduces noise and impairs the representation capacity of the model. We also observe that our proposed method, MVMI-FT, outperforms all other baselines, indicating the effectiveness of learning representations across feature and topology views. In particular, compared with the best results of MLP, our proposed method achieves improvement of $1.81\%$. For topology synthetic dataset (\textsc{Top. Syn. Data}), all models which can utilize information content from the topology view achieve an accuracy of $100\%$, but MLP only achieves an accuracy of $95.13$ $\pm$ $0.21\%$. 

In general, experimental results on synthetic datasets illustrate that for unsupervised graph representation learning, only focusing on certain view may limit the graph learning model’s representation capacity, and our proposed method can focus on feature and topology views to effectively learn node representations.
\begin{table*}[ht]
	\caption{Embedding comparison under the clustering protocol using Calinski-Harabaz (CH) index and Silhouette (SH) coefficient.}
	\scalebox{0.83}{
		\begin{tabular}{|c|cc|cc|cc|cc|cc|cc|}
			\toprule
			\multirow{2}{*}{\textsc{Methods}}&\multicolumn{2}{c|}{\textsc{Cora}} &
			\multicolumn{2}{c|}{\textsc{CiteSeer}} & \multicolumn{2}{c|}{\textsc{PubMed}} &
			\multicolumn{2}{c|}{\textsc{PubMedFull}} & \multicolumn{2}{c|}{\textsc{Ama. Computers}} & \multicolumn{2}{c|}{\textsc{Ama. Photo}}\\
			&CH&SH&CH&SH&CH&SH&CH&SH&CH&SH&CH&SH\\
			\midrule
			\midrule
			\textsc{Dgi}&85.932&0.0488&57.326&0.0139&1456.21&0.0615&973.036&0.0384&432.084&0.0036&453.981&0.0535\\
			\textsc{Cmvr}&120.174&0.0698&100.118&0.0542&1674.61&0.0754&1077.60&0.0588&487.985&0.0089&457.559&0.0595\\
			\textsc{Mvmi-Ft}&\textbf{126.725}&\textbf{0.0899}&\textbf{131.109}&\textbf{0.1176}&\textbf{1932.31}&\textbf{0.0827}&\textbf{1484.71}&\textbf{0.0635}&\textbf{603.278}&\textbf{0.0092}&\textbf{462.216}&\textbf{0.0861}\\
			\bottomrule
	\end{tabular}}
	\centering
	\label{tabv}
\end{table*}
\begin{figure}[t]
	\centering
	\includegraphics[width=1.01\columnwidth]{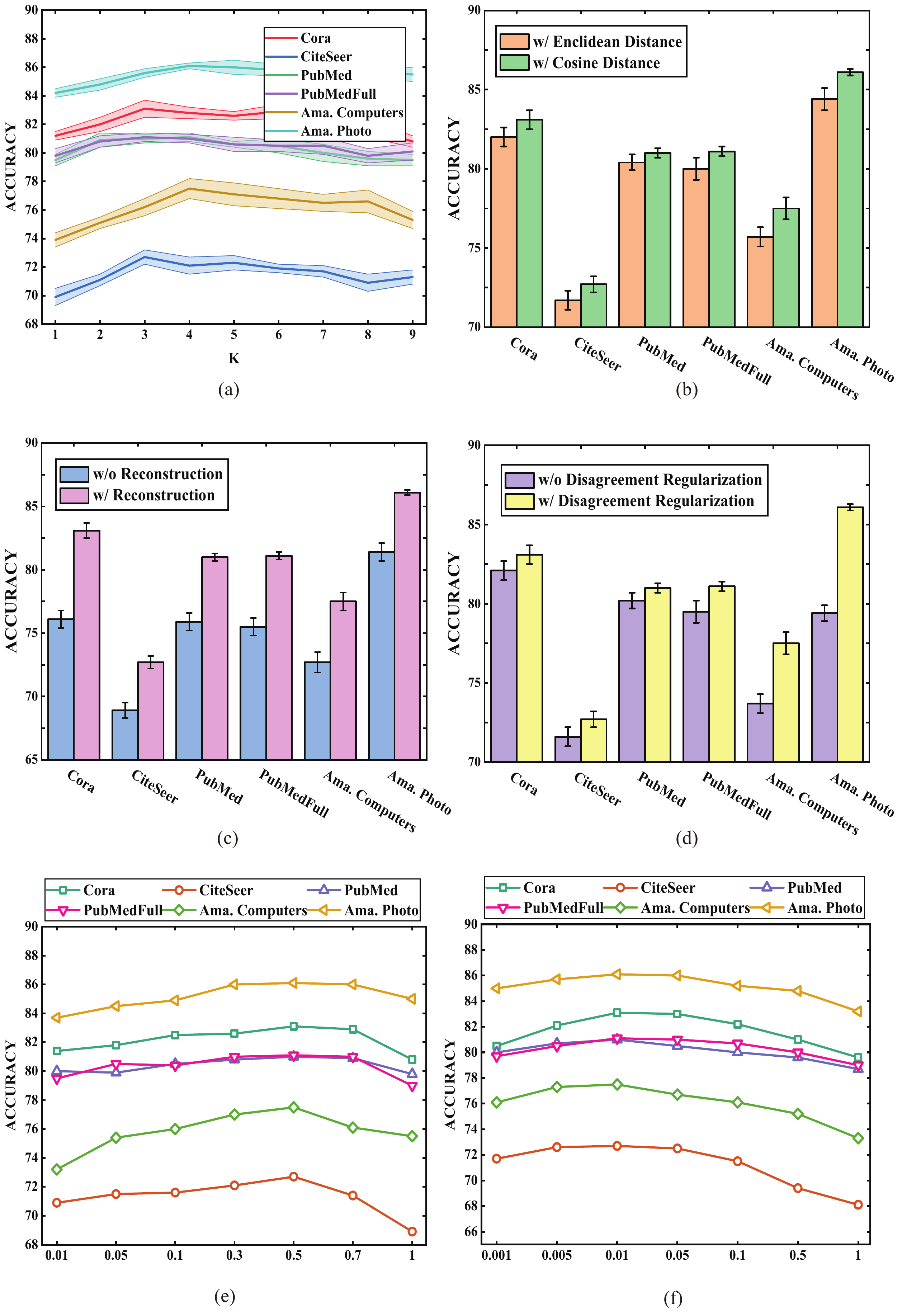}
	\caption{Ablation study on real-world datasets where (a) the effect of different number of $K$; (b) the effect of different distance metric in KNN search; (c) the effect of disagreement regularization; (d) the effect of reconstruction objective; (e) the effect of reconstruction module coefficient $\lambda_c$; (f) the effect of disagreement regularization coefficient $\lambda_d$.}
	\label{com}
\end{figure}
\subsection{Results on Real-World Datasets}
\subsubsection{Node classification results}
The node classification results on real-world datasets are shown in Table \ref{tabr}. Compared with the unsupervised methods, our proposed methods generally achieve the state-of-the-art classification performance on all datasets.  Compared with the supervised methods, our proposed method achieves comparable or even better performance under the unsupervised representation and linear evaluation protocol. To investigate the effect of different encoder for unsupervised graph representation learning, we utilize the SAGE encoder instead of GCN encoder in VGAE, CMVR, and our proposed MVMIFT model. Experimental results show that our method can utilize different graph encoders to enhance the model. To evaluate the effectiveness of multi-view representation, we develop the MI-FT model which only maximizes MI within each views. Compared with MI-FT, the MVMI-FT (MV and FT denote Multi-View and Feature-Topology respectively) consistently outperforms the MI-FT, indicating the effectiveness of multi-view representation. 
Note that CMVR achieves competitive results by utilizing the multi-view representation module across $1$-hop neighbors and Personalized PageRank diffusion views. Therefore, we also develop a variant, MVMI-FT-D (D denotes Diffusion), which fuses MVMI-FT with the Personalized PageRank diffusion operator. Considering the efficiency issue, we utilize the APPNP, which achieves linear computational complexity by approximating PageRank using power iteration \cite{appnp}, as the encoder in this model. Experimental results show that diffusion helps to further improve model performance, and MVMI-FT-D achieves the best classification accuracy on all datasets.

\subsubsection{Node clustering results}
The node clustering results on real-world datasets are shown in Table \ref{tabc}. From this table, we can observe that our proposed method, MVMI-FT, achieves state-of-the-art Normalized Mutual Information (NMI) and Adjusted Rand Index (ARI) scores compared with the previous methods, DGI and CMVR. Specifically, our method exhibits favourable improvement in clustering performance (the maximum gain of 18.1\% of the NMI and 23.4\% of the ARI scores for \textsc{PubMedFull} dataset) and its improvement are mainly driven by including graph feature topology, which demonstrate the effectiveness of our proposed method.  

\begin{figure*}[ht]
	\centering
	\includegraphics[width=1.9\columnwidth]{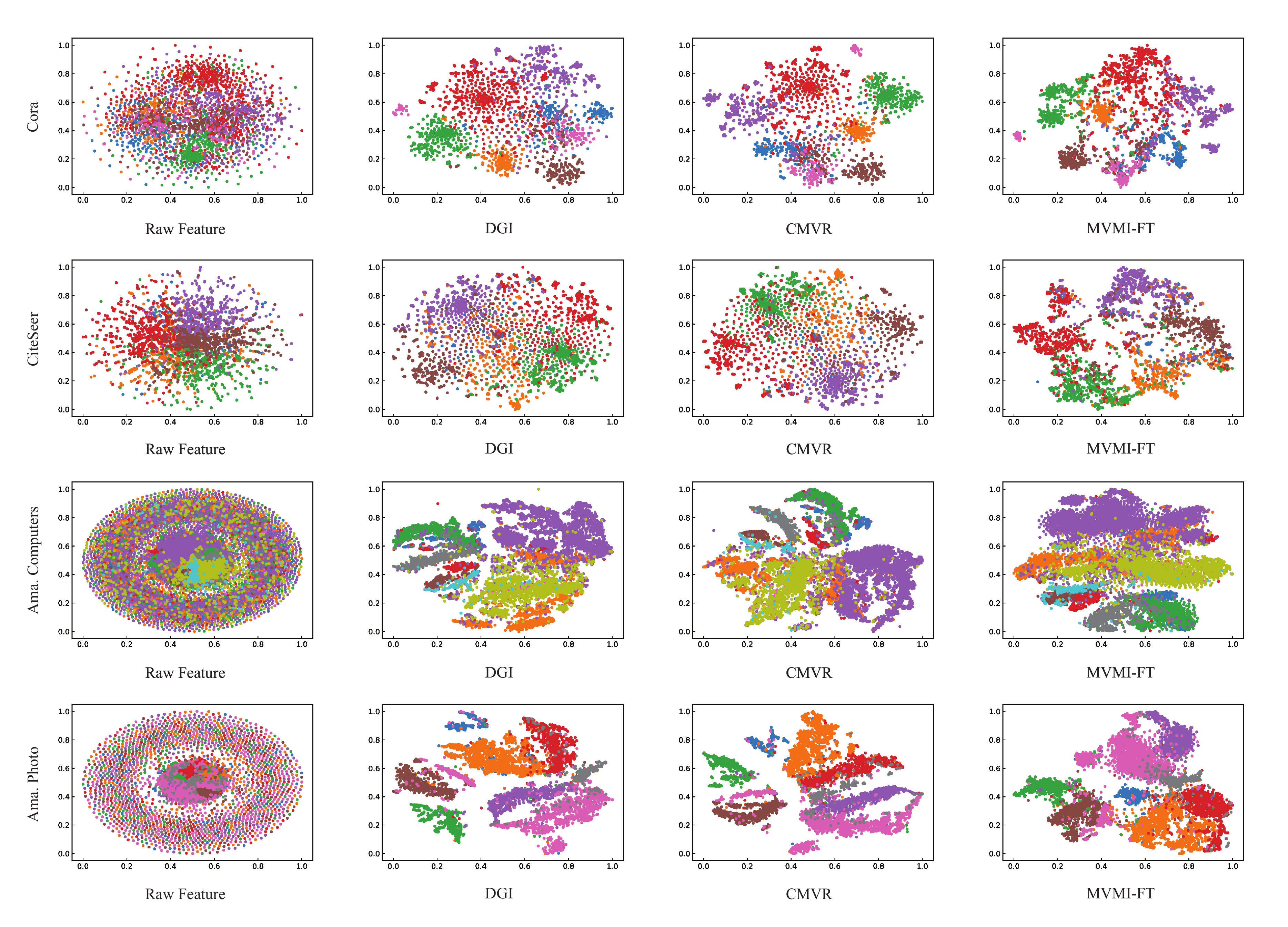}
	\caption{t-SNE visualization on \textsc{Cora}, \textsc{CiteSeer}, \textsc{Ama. Computers}, and \textsc{Ama. Photo} datasets. Different colors indicate the different labels.}
	\label{t_sne}
\end{figure*}

\subsection{Ablation Study}
In this subsection, we conduct experiments for the purpose of ablation study. Experimental results are shown in Figure \ref{com}.

\subsubsection{Effect of parameter K in $k$-Nearest Neighbor search}
To investigate the effect of parameter $K$ in $k$-nearest neighbor search, we utilize the proposed model, MVMI-FT, with respect to different KNN parameter $K = [1, 9]$. From Figure \ref{com} (a), we can observe that the classification performance on all datasets is increase first. With the increase of $K$, the performance stabilizes and then start to decrease. The probably reason is that larger $K$ may cause graph to be more denser which makes node representations indistinguishable. The model achieves the best performance on \textsc{Cora}, \textsc{CiteSeer}, \textsc{PubMed}, and \textsc{PubMedFull} when $K = 3$, \textsc{Ama. Computers} and \textsc{Ama. Photo} when $k=4$. Hence, we set the hyper parameter $K$ to be $3$ for \textsc{Cora}, \textsc{CiteSeer}, \textsc{PubMed}, and \textsc{PubMedFull} datasets, $4$ for Amazon datasets in our model. 

\subsubsection{Effect of different distance metric in $k$-Nearest Neighbor search}
To investigate the effect of different distance metric in  $k$-nearest neighbor search, we use the Cosine distance and Enclidean distance as similarity metric, respectively. The results are shown in Figure \ref{com} (b). From this figure, we can observe that using the cosine distance is better than the Enclidean distance for measuring node representation sample similarity in $k$-nearest neighbor search. Therefore, we set the similarity metric to be cosine distance in our model. 

\subsubsection{Effect of reconstruction loss}
To investigate the performance of model with or without reconstruction loss, we conduct the comparative experiments for all datasets using two models (w/o Reconstruction and w/  Reconstruction). The experimental results are shown in Figure \ref{com} (c). From this figure, we can clearly find that the model with the reconstruction loss generally performs better than without the reconstruction loss, indicating the necessity of the reconstruction for learning the common representations.

\subsubsection{Effect of disagreement regularization}
To investigate the performance of model with or without disagreement regularization $\mathcal{L}_d$, we compare these two models (w/o Disagreement Regularization and w/ Disagreement Regularization) on all datasets to validate the effectiveness of the constraints. Experimental results are shown in Figure \ref{com} (d). From this figure, we can observe that with disagreement regularization consistently yields better results, verifying the usefulness of the  disagreement regularization. 

\subsubsection{Effect of common representation objective coefficient}
To investigate the sensitivity of the common representation objective coefficient $\lambda_c$, we set $\lambda_c$ to be  $\{0.01, 0.05, 0.1, 0.3, 0.5, 0.7, 1\}$ and report the classification performance in Figure \ref{com} (e). For clarity, we do not report the error bar in this figure. From this figure, we can observe that with the increase of the common representation objective coefficient $\lambda_c$, the performance raise first and then start to decrease slowly. When $\lambda_c = 0.5$, model achieves best performance on all datasets. We therefore set the common representation objective coefficient $\lambda_c$ to be $0.5$ in our model.  
\subsubsection{Effect of disagreement regularization coefficient}
To investigate the sensitivity of the disagreement regularization coefficient $\lambda_d$, we conduct comparative experiments by using different coefficient settings ranging from $\{0.001, 0.005, 0.01, 0.05, 0.1, 0.5, 1\}$. The experimental results are shown in Figure \ref{com} (f). From this figure, we can find a similar observation with the experiment about common representation objective coefficient, \textit{i.e.}, with the increase of the disagreement regularization coefficient $\lambda_d$, the performance raise first and then start to decrease slowly. When $\lambda_d = 0.01$, model achieves best performance on all datasets. Hence, we set the disagreement regularization coefficient $\lambda_d$ to be $0.01$ in our model.    

\subsection{Quantitative Investigation}

To quantitatively investigate the embedding performance of our proposed model, we attempt to measure clustering quality by calculating the Calinski-Harabaz (CH) index \cite{calinski} and Silhouette (CH) coefficient \cite{silhouettes} score of the learned embedding representations using DGI, CMVR, and the proposed MVMI-FT. Note that different from applying $K$-means algorithm for clustering evaluation, we only evaluate the learned embedding representations in this experiment for the purpose of quantitatively investigation. We use built-in functions about these two metrics in the scikit-learn Python package with default settings for evaluation. The experimental results are shown in Table \ref{tabv}. From these results, we can observe that our method, MVMI-FT, consistently outperforms the previous method, DGI and CMVR, illustrating the effectiveness of our method for unsupervised graph representation learning.  

\subsection{Qualitative Visualization}

To qualitatively evaluate the embedding performance of the developed model, we use t-SNE \cite{tsne} to visualize the learned node embedding representations on the Cora, CiteSeer, Amazon Computers, and Amazon Photo datasets. The visualized results are displayed in Figure \ref{t_sne}. From this figure, we can observe that the distribution of plots learned by DGI and CMVR seems to be similar, and the embedding representations generated by our method, MVMI-FT, exhibit more discernible clusters than the raw feature, DGI and CMVR from a qualitative perspective, further indicating the effectiveness of our method for unsupervised graph representation learning.

\section{Conclusion} \label{s6}
While recent works only focus on the topology view for learning graph representations using mutual information maximization, we show that the information content coming from the feature view also plays an important role in unsupervised graph representation learning. Based on this observation, we propose a new approach to learn graph representations using mutual information maximization which consists of multi-view and common representation modules. To explicitly encourage the diversity between representations from same views, we introduce a disagreement regularization as objective constraint. Experimental results on synthetic and real-world datasets verify the effectiveness of our developed approach. our proposed method achieve new state-of-the-art results compared with the previous unsupervised methods, and achieve comparable or even better performance under the unsupervised representation and linear evaluation protocol compared with the previous supervised methods on real-world datasets. In the future, we are planning to explore feasible unsupervised graph representation approaches using mutual information maximization towards higher efficiency and larger scale.

\bibliographystyle{IEEEtran}
\bibliography{IEEEabrv,reference}

\end{document}